\documentclass{article}

\usepackage[final, nonatbib]{neurips_2021}

\usepackage[utf8]{inputenc} 
\usepackage[T1]{fontenc}    
\usepackage{hyperref}       
\usepackage{url}            
\usepackage{booktabs}       
\usepackage{amsfonts}       
\usepackage{nicefrac}       
\usepackage{microtype}      
\usepackage{xcolor}         
\usepackage{mathtools}      
\usepackage{wrapfig}
\usepackage[ruled]{algorithm2e}

\usepackage[maxnames=100]{biblatex}
\DeclareMathOperator{\defeq}{\stackrel{\text{def}}{=}}
\DeclareMathOperator*{\argmax}{arg\,max}
\SetKwComment{Comment}{/* }{ */}

\title{An Algorithmic Theory of Metacognition\\in Minds and Machines}

\author{%
  Rylan Schaeffer \\
  Department of Computer Science\\
  Stanford University\\
  \texttt{rschaef@stanford.edu} 
}

\begin{document}

\maketitle

\begin{abstract}
  Humans sometimes choose actions that they themselves can identify as sub-optimal, or wrong, even in the absence of additional information.  How is this possible? We present an algorithmic theory of metacognition based on a well-understood trade-off in reinforcement learning (RL) between value-based RL and policy-based RL. To the cognitive (neuro)science community, our theory answers the outstanding question of why information can be used for error detection but not for action selection. To the machine learning community, our proposed theory creates a novel interaction between the Actor and Critic in Actor-Critic agents and notes a novel connection between RL and Bayesian Optimization. We call our proposed agent the \textbf{Metacognitive Actor Critic (MAC)}. We conclude with showing how to create metacognition in machines by implementing a deep MAC and showing that it can detect (some of) its own suboptimal actions without external information or delay.
\end{abstract}

\section{Introduction}

Metacognition has a rich history in cognitive (neuro)science, but a relatively scarce history in machine learning. Why? One possible explanation is that a detailed, complete and persuasive theory of metacognition does not yet exist. Multiple theories have been put forward, but none are satisfactory for a variety of reasons (see Sec \ref{sec:theories_of_metacognition} Theories of Metacognition). Of particular interest is a recent and influential Bayesian computational theory \cite{fleming_self-evaluation_2017}, which posits that metacognition is ``'second-order' inference on a coupled but distinct decision system, computationally equivalent to inferring the performance of another actor." The theory is attractive due to its ability to explain a variety of behavioral data, but the theory is unsatisfying because it critically relies on information being available to the agent that the agent \textit{can use} for action evaluation but \textit{cannot use} for action selection. As one researcher wrote in private correspondence, ``[In this theory of metacognition], you have extra sources of information that you choose not to employ for making the decision itself. When should the left hand not tell the right hand what it knows?!?"

In this work, we provide an algorithmic theory of metacognition which posits that metacognition arises not from separate information, but from two separate learning algorithms with complementary strengths that are well-understood in reinforcement learning (RL). To the cognitive science community, our algorithmic theory explains how an agent can detect its own suboptimal action without external input or additional information, resolving the puzzle at the center of the aforementioned computational theory \cite{fleming_self-evaluation_2017}. To the RL community, our theory creates a novel interaction between the Actor and Critic, and creates a new connection between Actor-Critics in RL and acquisition functions/surrogate functions in Bayesian Optimization \cite{frazier_tutorial_2018}. We create a deep neural network implementation of our proposed \textbf{Metacognitive Actor Critic (MAC)} and demonstrate that it can detect a subset of its own suboptimal actions with no external information or delay.

\section{Background}

Because this paper is aimed at both the cognitive science and machine learning communities, we start with relatively extensive summaries of both fields. An experienced reader can skip to Sec. \ref{sec:results} Results.

\subsection{Metacognition}

One common approach to studying metacognition consists of tasking participants to complete a task while additionally asking for an implicit or explicit subjective evaluation of their task performance or conscious awareness. Here we summarize three puzzling well-reproduced experimental findings.

\subsubsection{Ubiquity of Hyper-Metacognitive Sensitivity}

\begin{wrapfigure}[22]{r}{0.45\textwidth}
  \begin{center}
    \includegraphics[width=0.43\textwidth]{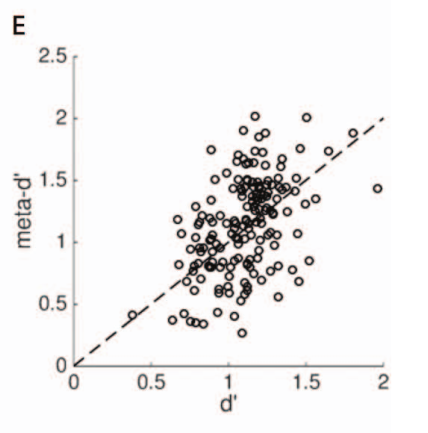}
  \end{center}
  \caption{\textbf{Meta-Cognitive Sensitivity.} Task performance ($d'$) should upper bound self-evaluation performance (meta-$d'$), but about half of participants display better evaluation than decision-making. Figure from \cite{fleming_self-evaluation_2017}}
  \label{fig:sensitivity}
\end{wrapfigure}

The first puzzling finding of metacognition is the ubiquity of a phenomenon termed ``hyper metacognitive sensitivity." Many perceptual decision-making tasks are 2-alternative forced choices (2AFC) tasks, in which a noisy stimulus is displayed and the participant needs to choose one of two mutually exclusive options (termed the ``Type 1" performance) before evaluating their own performance (termed the ``Type 2" performance). Under Signal Detection Theory, one can compute the signal-to-noise ratio available for the Type 1 performance: $d' \defeq \mu_s - \mu_n / \sqrt{0.5 * (\sigma_s^2 + \sigma_n^2)}$. We can define a similar quantity for Type 2 performance, which is termed meta-$d'$. The Type 1 performance $d'$ provides a theoretical upper bound on Type 2 performance meta-$d'$ \cite{galvin_type_2003}. However, a pooled analysis across studies shows that approximately half of participants have meta-$d'$ greater than $d'$; the implication is that participants have access to information that they use when judging whether their actions are correct, but not when selecting actions. If this is true, why does the brain not use that information to select better actions?

\subsubsection{Response-Locked Error Related Negativity Signal}

\begin{wrapfigure}[19]{r}{0.45\textwidth}
  \begin{center}
    \includegraphics[width=0.43\textwidth]{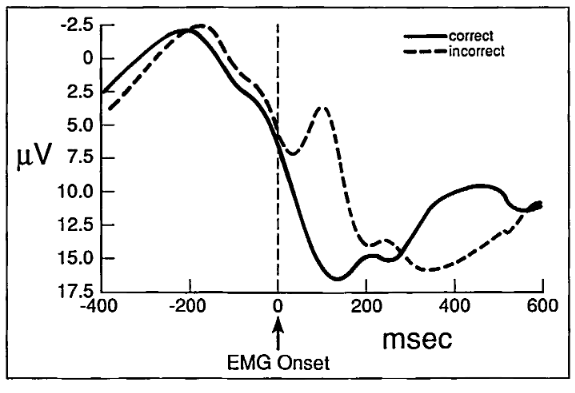}
  \end{center}
  \caption{\textbf{Response-Locked Error Related Negativity.} The Response-Locked ERN is an potential that appears following an incorrect action on a timescale too short to be driven by external information. Figure from \cite{gehring_neural_1993}.}
  \label{fig:ern}
\end{wrapfigure}

The second puzzling phenomenon of metacognition is the ability of humans to detect their own errors. In 1993, the so-called response-locked error related negativity (ERN) \cite{gehring_neural_1993} was discovered. The ERN is an event-related potential that peaks 70-110 ms following an incorrect action, that was initially observed following incorrect actions in a speeded choice reaction task (Fig. \ref{fig:ern}). The latency was too short for external information to drive such a difference, raising the question of how the brain knew a mistake had been made. Initial explanations suggested that the ERN was caused by a discrepancy between an efference copy of the executed motor command and the desired motor command, but several studies showed that the ERN still apears in passive observers \cite{}. Another study found that the ERN increases with subjective confidence, meaning the ERN has more complex expectations than mere binary correct / incorrect outcomes\cite{scheffers_performance_2000}. To further complicate matters, the ERN magnitude depends not only on whether an action is correct or incorrect, but on how early or late the actions were relative to the optimal action \cite{luu_medial_2000}, implying that the mechanism generating the ERN has expectations over both which actions should occur as well as when actions should occur; as the others puzzled, ``The amplitude of the ERN does not seem to be influenced by stimulus differences, nor does it seem to be involved in attempts to inhibit or correct erroneous responses or in making response selections." In our view, the question of what generates the ERN is an open one.

\subsubsection{Dissociability of Decision-Making and Confidence}

In general, the variability in participants' self-evaluation tracks their task performance \cite{fleming_self-evaluation_2017}. However, multiple experiments and interventions have shown that decision-making can be dissociated from confidence. This is the third puzzling phenomenon of metacognition we mention here. In one direction, interventions can affect confidence without affecting task performance. For instance, pharmacological manipulations with noradrenaline blockade increase the accuracy of self-evaluation without affecting task performance \cite{hauser_noradrenaline_2017}, and transcranial magnetic stimulation can decrease the accuracy of self-evaluation \cite{rounis_theta-burst_2010}. Age \cite{palmer_effects_2014} and lesions \cite{fleming_domain-specific_2014} can also affect decision confidence, without affecting task performance. In the other direction, interventions can affect task performance without affecting the accuracy of self-evaluation. The best examples are obsessive compulsive disorder (OCD) patients, who are frequently noted as subjectively knowing their habits are harmful but being unable to stop them. One 2017 study \cite{vaghi_compulsivity_2017} demonstrated that patients with OCD “show exaggerated action updating” (i.e. suboptimal action selection) and that  “Confidence in OCD is intact and reflects information that is not used to control action.” This evidence suggests that action selection and action evaluation are least partially dissociable.

\subsubsection{Theories of Metacognition}\label{sec:theories_of_metacognition}

Here, we briefly list some theories of metacognition and their criticisms. The Confidence-as-Bayesian-Posterior theory (\cite{meyniel_confidence_2015, pouget_confidence_2016}) posits that metacognitive confidence is the Bayesian posterior probability of choosing the correct action, a notion that does not generalize when an action cannot be defined as ``correct" and that also disagrees with experimental findings \cite{adler_limitations_2018}. Conflict Monitoring Theory (\cite{yeung_neural_2004}) suggests that error detection arises from multiple competing responses, and Predicted Response-Outcome theory (\cite{alexander_computational_2010, alexander_general_2014}) posits that error detection arises from discrepancies between predictions and outcomes, which is no different than ordinary RL. Perhaps most relevant to our paper is another Actor Critic (\cite{holroyd_mechanism_2005}), but that approach relies on the value difference between subsequent states $V(S_{n+1}) - V(S_n)$, which requires the agent to receive additional input to detect its own errors; in contrast, our MAC can detect its mistakes without additional input.

\subsection{Reinforcement Learning}

\subsubsection{Overview of Reinforcement Learning}

Reinforcement learning (RL) is a mathematical framework for studying how an agent in an environment can learn to select actions to maximize its cumulative reward. We consider an agent in an environment (formalized as a Markov Decision Process), and denote a single trajectory through the environment $\tau$ as the sequence of state ($S$), action ($A$) and reward ($R$) random variables:
\begin{equation}
\tau \, \defeq \, (S_1, A_1, R_1, S_2, ..., S_{N-1}, A_{N-1}, R_{N-1}, S_N)
\end{equation}

A trajectory $\tau$ through the environment is a random quantity whose probability depends on the environment's dynamics and the agent's actions. Letting $\theta$ denote the agent's parameters, the probability of a trajectory factorizes as:
\begin{align*}
    p(\tau; \theta) &= p(S_1) p(A_1|S_1; \theta) p(S_2, R_1| S_1, A_1)  ... p(A_{N-1}|S_{N-1}; \theta) p(S_N, R_{N-1}|S_{N-1}, A_{N-1}) \\
    &=\prod_n p(S_{n+1}, R_n | S_n, A_n) p(A_n | S_n; \theta)
\end{align*}

The total return over a trajectory is commonly defined either as the sum of rewards $G(\tau) \defeq \sum_n R_n$ or as the sum of the discounted rewards $G(\tau) \defeq \sum_n \gamma^{n-1} R_n$ for some discount factor $\gamma \in (0, 1)$. The agent then optimizes its parameters $\theta$ to select actions that in turn maximize its expected return:
\[\argmax_{\theta} \mathbb{E}_{p(\tau; \theta)} [G(\tau)] \]

The central question of RL is how should an agent approach this optimization problem. There are two families of approaches, called value-based RL and policy-based RL. 

\subsubsection{Value-Based Reinforcement Learning}

In value-based RL, the agent uses its parameters $\theta$ to predict the immediate reward and subsequent return of a particular state-action combination:
$$ Q(S, A; \theta) \defeq \mathbb{E}[G(\tau) | S, A]$$

To choose actions, the agent leverages the predicted value of each action. There are multiple ways of constructing a policy, for example the deterministic greedy policy:
\begin{align*}
    p(A_n | S_n; \theta) &= \begin{cases}
    1 & \text{if } A_n = \argmax_{A} Q(S_n, A; \theta)\\
    0 & \text{otherwise}
    \end{cases}
\end{align*}

The fundamental problem with value-based RL is that it fails in high-dimensional or continuous action spaces because finding the action which maximizes $Q(S_n, A; \theta)$ requires evaluating $Q(S_n, A; \theta)$ for every $A \in \mathcal{A}$, and as $|\mathcal{A}| \rightarrow \infty$, solving $\argmax_A Q(S_n, A;\theta)$ becomes infeasible. The intuition is that $Q(S_n, A;\theta)$ only tells the agent how good a state-action pair is, not which state-action pair is best; to determine the best state-action pair, the agent needs to query all possible state-action pairs, which becomes impractical if there are too many. Because challenging problems have high-dimensional (e.g. language modeling) or continuous (e.g. motor control) action spaces, value-based RL is rarely used by itself; most, if not all, notable RL triumphs select actions using policy-based RL.

\subsubsection{Policy-Based Reinforcement Learning}

In policy-based RL, rather than learning to predict the value of particular actions and then using those predicted values to choose actions, the agent instead uses its parameters $\theta$ to parameterize a probability distribution over the action space $\mathcal{A}$, and then samples actions from that distribution. %
$$A_n \sim p(A|S_N; \theta)$$

Because a distribution can be defined over a high-dimensional or continuous action space, policy-based RL does not suffer the same shortcoming as value-based RL. It does, however, have its own shortcoming: policy-based RL is significantly slower to learn than value-based RL, a shortcoming that can be ameliorated by pairing the two together, which we explain below.

\subsubsection{Policy Gradient Theorem}

In policy-based RL, how does the agent adjust its parameters to select actions that lead to a higher expected return? One ubiquitous approach is to compute the gradient of the expected return with respect to the agent's parameters and then perform gradient ascent. To compute the gradient of the expected return with respect to its parameters, the agent uses the \textit{policy gradient theorem}.
\begin{equation}
\begin{aligned}
    \nabla_{\theta} \mathbb{E}_{p(\tau; \theta)} [G(\tau)] &= \nabla_{\theta} \int G(\tau) p(\tau; \theta)  d\tau\\ 
    &= \int G(\tau) p(\tau; \theta ) \nabla_{\theta} \log p(\tau; \theta) d\tau\\ 
    &= \int G(\tau) p(\tau; \theta ) \nabla_{\theta} \sum_n \log p(A_n|S_n; \theta) d\tau\\ 
    &= \mathbb{E}_{p(\tau; \theta)} \Big[G(\tau) \nabla_{\theta} \sum_n \log p(A_n|S_n; \theta) \Big]
\end{aligned}
\end{equation}\label{eq:policy_grad_thm}

\subsubsection{How Value-Based RL Helps Policy-Based RL}

One significant problem with policy-based RL is that the return of a trajectory, $G(\tau)$, can vary wildly depending on the agent's actions and the environment's dynamics. This variance hampers the sample efficiency of policy-based RL because the fluctuating return is a noisy target to learn from. To address this problem, one popular approach is to use value-based RL to reduce the variance of $G(\tau)$ via a method known as \textbf{control variates} (called baselines in the RL literature). The idea is to replace the quantity we want to maximize, the return $G(\tau)$, with another quantity $\widetilde{G}(\tau)$ that has the same expected value but provably lower variance. To do this, we need two ingredients: a scalar $\beta \in \mathbb{R}$ and a scalar function $Q: \tau \rightarrow \mathbb{R}$. We define the modified return as:
\begin{equation*}
    \widetilde{G}(\tau) \defeq G(\tau) - \beta \big( Q(\tau) - \mathbb{E}[Q(\tau)] \big)
\end{equation*}

First, note that the modified return $\widetilde{G}(\tau)$ has the same expected value as the original return $G(\tau)$. For brevity, we suppress that all expectations are with respect to the trajectory distribution $p(\tau; \theta)$.
\begin{align*}
\mathbb{E} [\widetilde{G}(\tau)]
&= \mathbb{E}[G(\tau) - \beta (Q(\tau) - \mathbb{E} [Q(\tau)])]\\
&= \mathbb{E}[G(\tau)] - \beta(\mathbb{E} [Q(\tau)] - \mathbb{E} [Q(\tau)])\\
&= \mathbb{E}[G(\tau)]
\end{align*}

Second, by cleverly choosing $\beta$, the modified return $\widetilde{G}(\tau)$ has provably lower variance than $G(\tau)$. To achieve this, we consider the variance of $\widetilde{G}(\tau)$, differentiate with respect to $\beta$, set equal to zero and solve for optimal $\beta$ that minimizes the variance.
\begin{align*}
\mathbb{V}[\widetilde{G}(\tau)] &= \mathbb{V}[G(\tau) - \beta (Q(\tau) - \mathbb{E} [Q(\tau)])]\\
&= \mathbb{V}[G(\tau)] + \beta^2 \mathbb{V}[Q(\tau)] - 2 \beta Cov[G(\tau), Q(\tau)]\\
0 &= \frac{\partial}{\partial \beta} \mathbb{V}[\widetilde{G}(\tau)]\\
\beta^* &= \frac{Cov[G(\tau),Q(\tau)]}{\mathbb{V}[Q(\tau)]}
\end{align*}

Substituting $\beta^*$ back in, we find that the introduction of the control variate $Q(\tau)$ has guaranteed us that $\mathbb{V}_{p(\tau; \theta)}[\widetilde{G}(\tau)] \leq \mathbb{V}_{p(\tau; \theta)}[G(\tau)]$:

\begin{align*}
\frac{\mathbb{V}[\widetilde{G}(\tau)]}{\mathbb{V}[G(\tau)]}
&= \frac{\mathbb{V}[G(\tau)] + {\beta^*}^2 \mathbb{V}[Q(\tau)] - 2 {\beta^*} Cov[G(\tau), Q(\tau)]}{\mathbb{V}[G(\tau)]}\\
&= \frac{\mathbb{V}[G(\tau)] + \frac{Cov[G(\tau),Q(\tau)]^2}{\mathbb{V}[Q(\tau)]} - 2 \frac{Cov[G(\tau),Q(\tau)]^2}{\mathbb{V}[Q(\tau)])}}{\mathbb{V}[G(\tau)]}\\
&= \frac{\mathbb{V}[G(\tau)] - \frac{Cov[G(\tau),Q(\tau)]^2}{\mathbb{V}[Q(\tau)])}}{\mathbb{V}[G(\tau)]}\\
&= 1 - Corr[G(\tau),Q(\tau)]^2
\end{align*}

Since the square of the correlation must be non-negative, the variance of $\widetilde{G}(\tau)$ must be smaller than $G(\tau)$. Consequently, rather than using Eqn. \ref{eq:policy_grad_thm}, the agent can instead use:
\begin{equation*}
    \nabla_{\theta} \mathbb{E}_{p(\tau; \theta)} [G(\tau)] = \mathbb{E}_{p(\tau; \theta)} \Big[\widetilde{G}(\tau) \nabla_{\theta} \sum_n \log p(A_n|S_n; \theta) \Big]
\end{equation*}\label{eq:policy_grad_thm_with_ctrl_var}

The intuition behind the control variate $Q(\tau)$ is that the agent can reduce the variance of the learning signal since the variance between $G(\tau)$ and $Q(\tau)$ will be smaller than the variance between $G(\tau)$ and $0$. The control variate converges much more quickly than the policy because it is solving the easier problem of action evaluation than the harder problem of action selection. Agents that learn a policy via policy-based RL and learn a control variate via value-based RL are called \textbf{Actor-Critics}.

\section{Results}\label{sec:results}

\subsection{Algorithmic Theory of Metacognition}

To summarize the background, RL tells us that there is a tradeoff between value-based RL and policy-based RL that incentivizes agents to use both:

\begin{enumerate}
    \item In order to select actions environments with high-dimensional (e.g. language modeling) or continuous (e.g. motor control) action spaces, agents should use policy-based RL to learn a policy $p(A|S_n;\theta)$.
    \item In order to stabilize the learning of policy-based RL, agents should use a control variate $Q(\tau)$ that learns to predict the return $G(\tau)$.
\end{enumerate}

\begin{figure}[b!]
    \centering
    \includegraphics[width=0.8\textwidth]{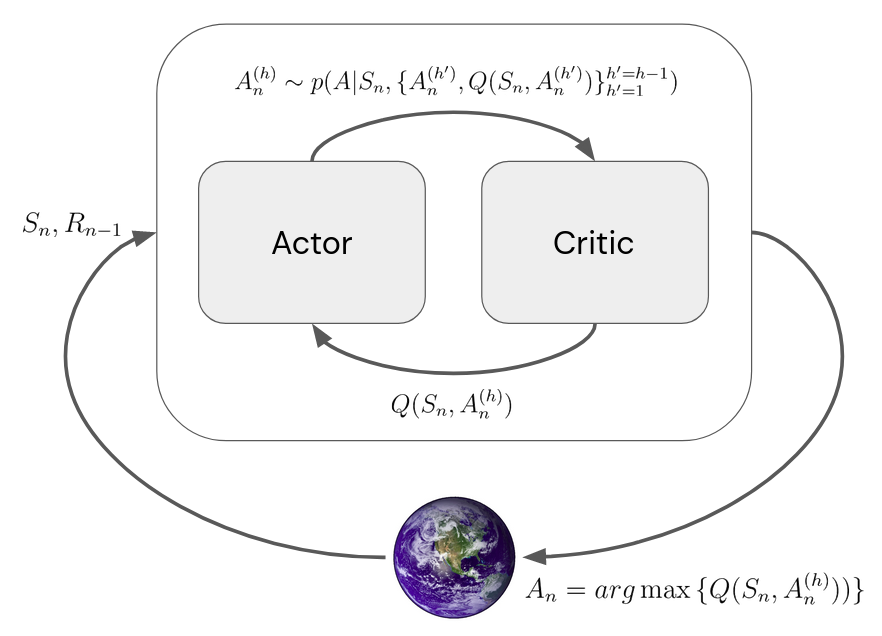}
    \caption{\textbf{Metacognitive Actor Critic.} On each step $n$ of the environment, the agent's Actor and Critic interact repeatedly. The Actor iteratively samples hypothetical actions $A_n^{(h)}$ and queries the Critic for their predicted values $Q(S_n, A_n^{(h)})$. The Actor uses the previous hypothetical actions and their predicted values to construct a new policy from which to sample the next hypothetical action $A_n^{(h+1)})$. This iterative process repeats until the agent is satisfied or until time pressure forces it to act. The agent's real action $A_n$ is chosen from the hypothetical actions e.g. greedily with $\argmax$.}
    \label{fig:mac}
\end{figure}

In the RL community, the control variate $Q(\tau)$ serves no purpose other than to help the policy learn faster. Here, we propose a new interaction between the two that we call the \textbf{Metacognitive Actor Critic (MAC)} (Fig. \ref{fig:mac}; Code \ref{alg:mac}). We first define how the MAC interacts with its environment and then explain how it creates behavior and neural signatures that cognitive neuroscience calls metacognition.

On each step of the environment, like any RL agent, the MAC receives the state $S_n$ and the reward $R_{n-1}$. The Critic immediately calculates $V(S_n)$. The Actor constructs its first policy and samples a hypothetical action $A_n^{(1)} \sim p(A| S_n)$. The Actor passes this hypothetical action to the Critic, which in turn predicts the hypothetical action's value $Q(S_n, A_n^{(1)})$. The Actor then constructs its second policy, conditioned on $A_n^{(1)}$ and $Q(S_n, A_n^{(1)})$, that it then uses to sample a new hypothetical action $A_n^{(2)} \sim p(A|S_n, \{ A_n^{(1)}, Q(S_n, A_n^{(1)})\})$. This process repeats, with the Actor sampling hypothetical actions and the Critic evaluating the hypothetical actions, until either the agent is satisfied or until time pressure forces it to act. The agent then chooses its real action from the set of hypothetical actions e.g. choosing the hypothetical action with the highest predicted value $A_n \defeq \argmax \Big\{ Q(S_n, A_n^{(h})) \Big\}$.

Returning to metacognition, how can the MAC detect its own errors? Suppose the MAC's Actor samples its first hypothetical action $A_n^{(1)}$ and sends the hypothetical action to the MAC's Critic. The Critic evaluates $Q(S_n, A_n^{(1)})$ and discovers that $Q(S_n, A_n^{(1)}) - V(S_n) < 0$, implying that $A_n^{(1)}$ is suboptimal. But suppose that the MAC is under time pressure, and while the Critic was doing its evaluation, the Actor chose hypothetical action $A_n^{(h)}$ as its real action $A_n$ and submitted that to the environment. Before any new external information can be received, the MAC can detect that the action was suboptimal. By having the Actor and Critic interact in this fashion, the MAC can detect its own erroneous actions and potentially sample better actions (if time permits).

The MAC has a clear correspondence with Bayesian optimization \cite{frazier_tutorial_2018}, with the Critic serving as the surrogate function and the Actor serving as the acquisition function deciding where to sample next. To the best of our knowledge, this relationship has not been previously noted or explored.

\bigskip

\begin{algorithm}
\caption{Metacognitive Actor Critic (MAC)}\label{alg:mac}
\For{\text{Environment Step} $n = 1, 2,  ..., N$}{

  Environment sends $S_n, R_{n-1}$ to MAC.
  
  MAC computes $V(S_n)$

  MAC initializes $\text{Hypothetical Actions: } \mathcal{A}_n \gets \{\}$
  
  MAC initializes $\text{Hypothetical Actions' Values: } \mathcal{Q}_n \gets \{\}$
  
  \For{\text{Hypothetical Evaluation} $h = 1, 2,  ..., H$}{
    
    MAC's Actor constructs a policy: $p(A|S_n, \mathcal{A}_n, \mathcal{Q}_n)$
    
    MAC's Actor samples a hypothetical action: $A_n^{(h)} \sim p(A|S_n, \mathcal{A}_n, \mathcal{Q}_n)$
    
    MAC's Critic evaluates the hypothetical action: $Q(S_n, A_n^{(h)})$
    
    MAC adds hypothetical action to set of hypothetical actions: $\mathcal{A}_n \gets \mathcal{A}_n \cup \{ A_n^{(h)} \}$
    
    MAC adds hypothetical action's value to set of hypothetical actions' values:
    $\mathcal{Q}_n \gets \mathcal{Q}_n \cup  \{ Q(S_n, A_n^{(h)}) \}$
  }
  
  MAC chooses real action from hypothetical actions e.g. $A_n \defeq \argmax \Big\{ Q(S_n, A_n^{(h})) \Big\}$
  
  MAC sends real action $A_n$ to Environment
}
\end{algorithm}

\subsection{Implementing the Metacognitive Actor Critic (MAC)}


\begin{wrapfigure}[18]{r}{0.5\textwidth}
  \begin{center}
    \includegraphics[width=0.47\textwidth]{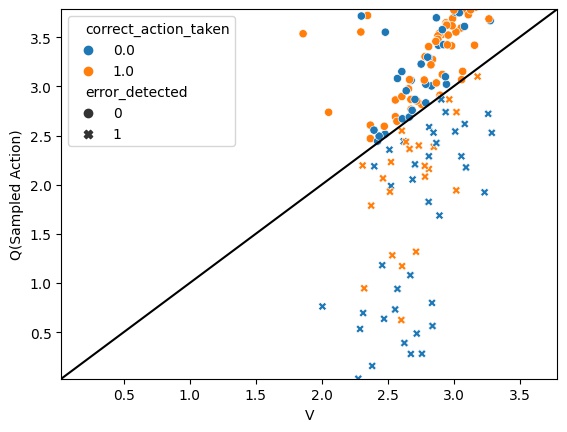}
  \end{center}
  \caption{\textbf{The MAC's Critic detects incorrect actions} whenever $Q(S_n, A_n) - V(S_n) < 0$.}
  \label{fig:mac_behav_1}
\end{wrapfigure}

We trained an LSTM-based \cite{hochreiter_long_1997} MAC agent to perform a psychophysics-inspired speeded-response two-alternative forced-choice (2AFC) task. On each step, the agent receives a sequence of paired noisy sensory inputs $(o_n^{Left}, o_n^{Right})$, sampled i.i.d. from two univariate Gaussians, one with mean $>0$ (signal) and the other with mean $=0$ (noise). The agent receives no inputs after the initial stimulus concludes. The agent outputs two quantities: first, the agent makes a binary decision as to which side (left or right) it thinks the signal appears, and second, the agent reports how confident it is in its decision. We assume the agent is under time pressure to act, and thus can only sample a single action. Our architecture uses three separate LSTMs that respectively output $V(S_n), p(A|S_n; \theta)$ and $Q(S_n, A_n)$; the action $A_n$ is sampled from the output of the $p(A|S_n)$ LSTM, and subsequently fed into the $Q(S_n, A_n)$ LSTM as input. The action $A_n$ is the agent's decision, and the confidence is defined as $Q(S_n, A_n) - V(S_n)$.

\begin{figure}
\begin{minipage}{0.5\textwidth}
        \centering
        \includegraphics[width=0.9\linewidth]{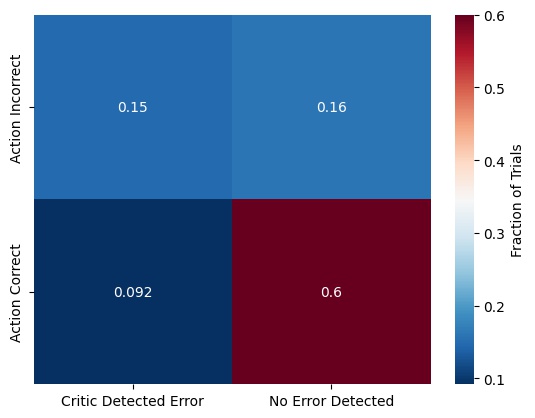}
    \end{minipage}%
    \begin{minipage}{0.5\textwidth}
        \centering
        \includegraphics[width=0.9\linewidth]{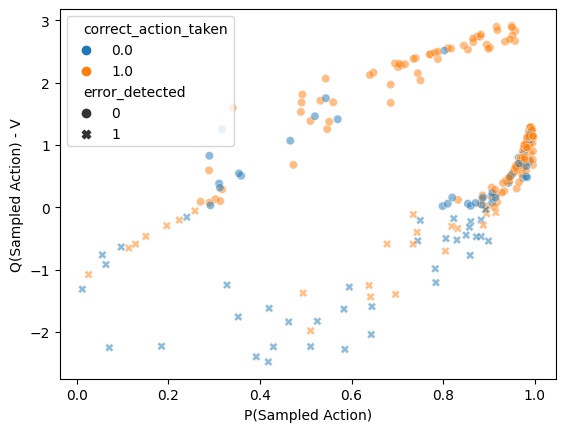}
    \end{minipage}
\caption{(A) The MAC detects approximately half of its erroneous decisions. (B) The MAC detects suboptimal action sampled from the policy, even actions sampled with high probability.}
\label{fig:mac_behav_2}
\end{figure}

We titrated the signal mean means till the Actor reached a respectable level of performance: choosing the correct side on 69\% of trials. By defining an error as whenever $Q(S_n, A_n) - V(S_n) < 0$, we found that the Critic is capable of recognizing a significant fraction of actions sampled by the Actor as incorrect (Fig. \ref{fig:mac_behav_1}). Investigating further, we found that 62.6\% (0.15/(0.15 + 0.092)) of trials in which the Critic detected an erroneous action were indeed incorrect trials (Fig. \ref{fig:mac_behav_2}A), meaning the Critic detects suboptimal actions better than chance. We also found that the Critic catches 48\% of the incorrect actions (Fig. \ref{fig:mac_behav_2}A). This means the MAC is capable of detecting its incorrect decisions and reporting an appropriate lack of confidence in the decision, without any additional external information. When we compared the error detection against the Actor's probability of sampling the chosen action, we found that many of the erroneous actions were sampled with high probability (>90\%), showing that the Critic has learnt information that the Actor has not yet learnt.

\section{Conclusion}

\section{Future Directions}

\subsection{Cognitive Neuroscience}

The immediate next step is to show that the MAC is capable of explaining the dissociability of action selection and action evaluation. By disrupting the normal functioning (e.g. injecting noise) of either the Actor or Critic, the MAC's decision-making or its confidence should be affected, but not both; this would be consistent with age, lesion, pharmacological and TMS studies. An additional step would be to show the MAC produces a neural signature akin to the ERN; supposing that the ERN is $Q(S_n, A_n^{(h)}) - V(S_n)$, the MAC would explain why OCD patients have increased ERNs and why the ERN amplitude correlates with the severity of symptoms: the addled Actor selects actions that the Critic recognizes as suboptimal, but the Actor refuses to choose differently.

\subsection{Machine Learning}

The next steps in RL are to show that the MAC outperforms vanilla actor critics in simple (e.g. grid world) environments. Subsequent engineering work should scale the agent up to challenging RL benchmarks and theoretical work should leverage tools from Bayesian optimization to guide how the MAC's Actor should adaptive sample actions in response to the insights provided by the Critic.

\clearpage

\printbibliography

\end{document}